# Representing and Solving Asymmetric Bayesian Decision Problems

**Thomas D. Nielsen**　　**Finn V. Jensen**
Department of Computer Science
Aalborg University
Fredrik Bajers Vej 7C, DK-9220 Aalborg Ø, Denmark

## Abstract

This paper deals with the representation and solution of asymmetric Bayesian decision problems. We present a formal framework, termed *asymmetric influence diagrams*, that is based on the influence diagram and allows an efficient representation of asymmetric decision problems. As opposed to existing frameworks, the asymmetric influence diagram primarily encodes asymmetry at the qualitative level and it can therefore be read directly from the model.

We give an algorithm for solving asymmetric influence diagrams. The algorithm initially decomposes the asymmetric decision problem into a structure of symmetric subproblems organized as a tree. A solution to the decision problem can then be found by propagating from the leaves towards the root using existing evaluation methods to solve the subproblems.

## 1  INTRODUCTION

The power of an influence diagram, both as an analysis tool and a communication tool, lies in its ability to concisely and precisely describe the structure of a decision problem[Smith et al., 1993]. However, influence diagrams can not efficiently represent the so-called asymmetric decision problems; decision problems are usually asymmetric in the sense that the set of possible outcomes of a chance variable may vary depending on the conditioning states, and the set of legitimate decision options of a decision variable may vary depending on the different information states [Qi et al., 1994].

Various frameworks have been proposed as alternatives to the influence diagram when dealing with asymmetric decision problems. [Covaliu and Oliver, 1995] extends the influence diagram with another diagram, termed a *sequential decision diagram*, which describes the asymmetric structure of the problem as complementary to the influence diagram which is used for specifying the probability model. [Smith et al., 1993] introduces the notion of *distribution trees* within the framework of influence diagrams. The use of distribution trees allows the possible outcomes of an observation to be specified, as well as the legitimate decision options of a decision variable. However, as the distribution trees are not part of the influence diagram, the structure of the decision problem can not be deduced directly from the model. Moreover, the sequence of decisions and observations is predetermined, i.e., previous observations and decisions can not influence the temporal order of future observations and decisions. Finally, distribution trees have a tendency of creating large conditionals during the evaluation since they encode both numeric information and information about asymmetry. To overcome this problem [Shenoy, 2000] presents the *asymmetric valuation network* as an extension of the *valuation network* for modelling symmetric decision problems[Shenoy, 1992]. The asymmetric valuation network uses *indicator functions* to encode asymmetry, thereby separating it from the numeric information. However, asymmetry is still not represented directly in the model and, as in [Smith et al., 1993], the sequence of observations and decisions is predetermined.[1]

In this paper we present the *asymmetric influence diagram* which is a framework for representing asymmetric decision problems. The asymmetric influence diagram is based on the *partial influence diagram*[Nielsen and Jensen, 1999b], and encodes structural asymmetry at the qualitative level; structural asymmetry has to do with the occurrence of variables in different scenarios as opposed to functional asymmetry which has to do with the possible out-comes/decision options of the variables. As a modelling language, the syntactical rules of the asymmetric influence diagram allow decision problems to be described in an easy and concise manner. Furthermore, its semantic specification supports an efficient evaluation algorithm.

---

[1] Further details and comparisons of these methods can be found in [Bielza and Shenoy, 1999].

An outline of this paper is as follows. In Section 2 we describe the partial influence diagram, together with the terms and notation used throughout this paper. In Section 3 we formally introduce the asymmetric influence diagram and illustrate this framework by modelling a highly asymmetric decision problem termed "the dating problem". Finally, in Section 4 we present an algorithm for solving asymmetric influence diagrams.

## 2 PRELIMINARIES

The partial influence diagram (PID) was defined in [Nielsen and Jensen, 1999b] as an influence diagram (ID) with only a partial temporal order over the decision nodes. That is, a PID is a directed acyclic graph $I = (\mathcal{U}, \mathcal{E})$, where the nodes $\mathcal{U}$ can be partitioned into three disjoint subsets; *chance nodes* $\mathcal{U}_C$, *decision nodes* $\mathcal{U}_D$ and *value nodes* $\mathcal{U}_V$. The chance nodes (drawn as circles) correspond to *chance variables*, and represent events which are not under the direct control of the decision maker. The decision nodes (drawn as squares) correspond to *decision variables* and represent actions under the direct control of the decision maker. We will use the concept of node and variable interchangeably if this does not introduce any inconsistency, and we assume that no *barren nodes* are specified by the PID since they have no impact on the decisions.[2]

With each chance variable and decision variable X we associate a finite discrete *state space* $W_X$ which denotes the set of possible outcomes/decision options for X. For a set $\mathcal{U}'$ of variables we define the state space as $W_{\mathcal{U}'} = \times \{W_X | X \in \mathcal{U}'\}$.

The set of value nodes (drawn as diamonds) defines a set of *utility potentials* with the restriction that value nodes have no descendants. Each utility potential indicates the local utility for a given configuration of the variables in its *domain*; the domain of a utility potential $\psi_X$, for a value node X, is denoted $\text{dom}(\psi_X) = \Pi_X$, where $\Pi_X$ is the immediate predecessors of X. The total utility is the sum or the product of the local utilities (see [Tatman and Shachter, 1990]); in the remainder of this paper we assume that the total utility is the sum of the local utilities.

---

[2] A chance node or a decision node is said to be barren if it does not precede any other node, or if all its descendants are barren.

The uncertainty associated with a variable $X \in \mathcal{U}_C$ is represented by a *conditional probability potential* $\phi_X = P(X|\Pi_X) : W_{X \cup \Pi_X} \to [0;1]$. The *domain* of a conditional probability potential $\phi_X$ is denoted $\text{dom}(\phi_X) = \{X\} \cup \Pi_X$.

The arcs in a PID can be partitioned into three disjoint subsets, corresponding to the type of node they go into. Arcs into value nodes represent functional dependencies by indicating the domain of the associated utility potential. Arcs into chance nodes, denoted *dependency arcs*, represent probabilistic dependencies, whereas arcs into decision nodes, denoted *informational arcs*, imply information precedence; if $D \in \mathcal{U}_D$ and $(X, D) \in \mathcal{E}$ then the state of X is known when decision D is made.

The set of informational arcs induces a *partial order* $\prec$ on $\mathcal{U}_C \cup \mathcal{U}_D$ as defined by the transitive closure of the following relation:

- $Y \prec D_i$, if $(Y, D_i)$ is a directed arc in I ($D_i \in \mathcal{U}_D$).

- $D_i \prec Y$, if $(D_i, X_1, X_2, \ldots, X_m, Y)$ is a directed path in I ($Y \in \mathcal{U}_C \cup \mathcal{U}_D$ and $D_i \in \mathcal{U}_D$).

- $D_i \prec A$, if $A \not\prec D_j$ for all $D_j \in \mathcal{U}_D$ ($A \in \mathcal{U}_C$ and $D_i \in \mathcal{U}_D$).

- $D_i \prec A$, if $A \not\prec D_i$ and $\exists D_j \in \mathcal{U}_D$ s.t. $D_i \prec D_j$ and $A \prec D_j$ ($A \in \mathcal{U}_C$ and $D_i \in \mathcal{U}_D$).

In what follows we say that two different nodes X and Y are *incompatible* if $X \not\prec Y$ and $Y \not\prec X$.

We define a *realization* of a PID I as an attachment of potentials to the appropriate variables in I, i.e., the chance nodes are associated with conditional probability potentials and the value nodes are associated with utility potentials. So, a realization specifies the quantitative part of the model whereas the PID constitutes the qualitative part.

Evaluating a PID amounts to computing a strategy for the decisions involved. A strategy can be seen as a prescription of responses to earlier observations and decisions, and it is usually performed according to the *maximum expected utility principle*; the maximum expected utility principle states that we shall always choose a decision option that maximizes the expected utility. However, the strategy for a decision variable may depend on the variables observed thus, we define an *admissible total order* for a PID I to describe the relative temporal order of incompatible variables: an admissible total order is a bijection $\beta : \mathcal{U}_D \cup \mathcal{U}_C \leftrightarrow \{1, 2, \ldots, |\mathcal{U}_D \cup \mathcal{U}_C|\}$ s.t. if $X \prec Y$ then $\beta(X) < \beta(Y)$, where $\prec$ is the partial order induced by I. In what follows $<$ will be used to denote the total ordering $\beta$ s.t. $X < Y$ if $\beta(X) < \beta(Y)$. Notice, that an admissible total ordering of a PID I implies that I can be seen as an ID.

Given an admissible total ordering $<$, we define an *admissible strategy relative to* $<$ as a set of functions $\Delta^< = \{\delta_D^\lessgtr | D \in \mathcal{U}_D\}$, where $\delta_D^\lessgtr$ is a decision function given by:

$$\delta_D^\lessgtr : W_{\text{pred}(D)^<} \to W_D,$$

and $\text{pred}(D)^< = \{X|X < D\}$ (the index $<$ in $\text{pred}(D)^<$ will be omitted if this does not introduce any confusion). Given a realization of a PID I, we term an admissible strategy relative to $<$, an *admissible optimal strategy relative to* $<$ if the strategy maximizes the expected utility for I; two admissible optimal strategies are said to be *identical* if they yield the same expected utility. A decision function $\delta_D^\lessgtr$, contained in an admissible optimal strategy relative to $<$, is said to be an *optimal strategy for* D *relative to* $<$. Note that an optimal strategy for a decision variable D relative to $<$ does not necessarily depend on all the variables observed. Hence, we say that an observed variable X is *required* for D w.r.t. $<$ if there is a realization of I s.t. the optimal strategy for D relative to $<$ is a non-constant function over X. By this we mean that there exists a configuration $\bar{y}$ over $\text{dom}(\delta_D^\lessgtr)\setminus\{X\}$ and two states $x_1$ and $x_2$ of X s.t. $\delta_D^\lessgtr(x_1, \bar{y}) \neq \delta_D^\lessgtr(x_2, \bar{y})$.

**Definition 1.** A realization of a PID I is said to *define a decision problem* if all admissible optimal strategies for I are identical. A PID is said to *define a decision problem* if all its realizations define a decision problem.

The above definition characterizes the class of PIDs which can be considered welldefined since the set of admissible total orderings for a PID I corresponds to the set of legal elimination sequences for I. However, it also conveys the problem of only having a partial temporal ordering of the decision variables; the relative temporal order of a chance variable (eliminated by summation) and a decision variable (eliminated by maximization) may vary under different admissible orderings and summation and maximization does not in general commute. So, in order to determine whether or not a PID defines a decision problem we introduce the notion of a *significant chance variable*.

**Definition 2.** Let I be a PID and let A be a chance variable incompatible with a decision variable D in I. Then A is said to be *significant* for D if there is a realization and an admissible total order $<$ for I s.t. :

- A occurs immediately before D under $<$.
- The optimal strategy for D relative to $<$ is different from the one achieved by permuting A and D in $<$.

Based on the above definition we have the following theorem which characterizes the constraints necessary and sufficient for a PID to define a decision problem.

**Theorem 1** ([Nielsen and Jensen, 1999b]). The PID I defines a decision problem if and only if for each decision variable D there does not exist a chance variable A significant for D.

See [Nielsen and Jensen, 1999b] for a structural characterization of the chance variables being significant for a given decision variable.

## 3 ASYMMETRIC INFLUENCE DIAGRAMS

[Qi et al., 1994] states that decision problems are usually asymmetric in the sense that the set of possible outcomes of a chance variable may vary depending on the conditioning states, and the set of legitimate decision options of a decision variable may vary depending on the different information states. Equivalently, [Bielza and Shenoy, 1999] characterizes a decision problem as being asymmetric if, in its decision tree representation, the number of scenarios is less than the cardinality of the Cartesian product of the state spaces of all chance and decision variables. However, both of these characterizations fail to recognize decision problems in which the relative temporal order of two variables vary w.r.t. to previous observations and decisions; this is for example very common for troubleshooting problems. Thus, we define an asymmetric decision problem as follows:

**Definition 3.** A decision problem is said to be *asymmetric* if, in its decision tree representation, either:

- the number of scenarios is less than the cardinality of the Cartesian product of the state spaces of all chance and decision variables or
- there exists two scenarios in which the relative temporal order of two variables differ.

In order to deal with such asymmetric decision problems we introduce the asymmetric influence diagram (AID). An AID is a labeled directed graph I = $(\mathcal{U}, \mathcal{E}, \mathcal{F})$, where the nodes $\mathcal{U}$ can be partitioned into four disjoint subsets; *test-decision nodes* ($\mathcal{U}_T$), *action-decision nodes* ($\mathcal{U}_A$), chance nodes ($\mathcal{U}_C$) and value nodes ($\mathcal{U}_V$); we will sometimes omit the distinction between test-decisions and action-decisions by simply referring to a node in $\mathcal{U}_D = \mathcal{U}_T \cup \mathcal{U}_A$ as a decision node.

The chance nodes and value nodes are similar to the chance nodes and value nodes in a PID. The decision nodes correspond to decision variables and represent actions under the direct control of the decision maker. A test-decision (drawn as a triangle) is a decision to look for more evidence, whereas an action-decision (drawn as a rectangle) is a decision to change the state of the world.

The arcs $\mathcal{E}$ in an AID can be partitioned into four disjoint subsets. An arc into a value node or a chance node is semanticly defined as in the PID framework if, in case of the latter, it does not emanate from a test-decision node. Arcs into decision nodes, termed informational arcs, imply a possible information precedence; if there is an arc from a node X to a decision node D then the state of X *may* be known when decision D is made. This redefinition is needed since we deal with asymmetric decision problems, i.e., the set of variables observed immediately before decision D is taken may dependent on previous decisions and observations.

If there exists an arc, termed a *test arc*, from a test-decision node D to a chance node X, then the state of D determines whether or not X is eventually observed; having an arc from a test-decision node to a chance node represents a logical relation and does not imply probabilistic dependence. Note that in the trivial case, where X is observed no matter the state of D, the arc (D, X) implies information precedence only. If there exists an arc (D, D') from a test-decision node D to another decision node D', then (D, D') implies information precedence ((D, D') is an informational arc) however, (D, D') is termed a test arc if the state of D determines whether or not D' is eventually decided upon; whether we are referring to an informational arc or a test arc is conveyed by the *label* associated with D'; in the remainder of this paper we let $\bar{I}$ denote the graph obtained from I by removing all test arcs and informational arcs.

The asymmetry of a decision problem is graphically represented in the AID by a set of *restriction arcs* and by a set of labels $\mathcal{F}$. The set of restriction arcs (drawn as dashed arcs) is a subset of the informational arcs. A restriction arc (X, D) indicates that the set of legitimate decision options for D may vary depending on the state of X, in which case we say that X is *restrictive* w.r.t. D (or X is restricting D). The set of labels $\mathcal{F}$ is associated with a subset of the nodes and informational arcs. A label specifies under which conditions the associated node or informational arc occurs in the decision problem. The following rules informally summarize the semantics of labels when specifying asymmetry in the AID; in the remainder of this section they will be referred to as rule (i)-(iii), respectively:

i) Let X be a node labelled with $f_X$ and let $\mathcal{Y}$ be the variables observed before X is observed (or decided upon). If $f_X$ is unsatisfied w.r.t. the state configuration $\mathcal{Y} = \bar{y}$ observed, then X is not included in the scenario.

ii) Let (X, D) be an informational arc labeled with $f_{(X,D)}$ and let $\mathcal{Y}$ be the variables observed before X is observed (or decided upon). If $f_{(X,D)}$ is unsatisfied w.r.t. the state configuration $\mathcal{Y} = \bar{y}$ observed, then (X, D) is not included in the scenario.

iii) If there exists a directed path from a node X to a node Y in $\bar{I}$, then whenever X is not included in the scenario Y is not included in the scenario either.

Based on the rules above we require that if the label of a node Z is a function of a node X, then there must exist an arc from X to Z (See Figure 1).

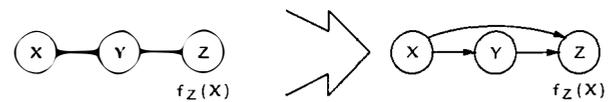

Figure 1: There must exist a directed arc from X to Z since $f_Z$ is a function of X.

**Example 1. [The dating problem]** Joe has to decide whether or not to ask a girl he has recently met on a date. If Joe decides not to ask her out he can choose either to stay at home and watch TV or visit a night club; before taking that decision Joe observes what programs will be on TV that night. The pleasure of staying at home is influenced by his liking of the program watched, whereas the pleasure of going to a night club is dependent on the comfort of going to that night club and the entrance fee; comfort is dependent on whether Joe likes the night club and whether he meets any friends there.

If Joe decides to ask her out her response will depend on her feelings towards him. If she declines the date, Joe can decide to go to a night club or stay at home and watch TV; we assume that the two "staying at home scenarios" are the same. If she accepts to go on a date with him, Joe will ask her whether she wants to go to a restaurant or to the movies. The choice of movie (decided by Joe) may influence her mood which in turn may influence Joe's satisfaction concerning the evening. Similarly, the choice of menu (decided by Joe) might influence Joe's satisfaction.

This decision problem is represented by the AID depicted in Figure 2. The variable *Date?* is represented as a test-decision since it has no impact on the value of *Accept?*. In the evaluation of whether or not to ask for a date, the distribution of *Accept?* is relevant and *Accept?* is therefore always part of the entire decision problem.

The decision *Night Club?* is decided upon if Joe ini-tially chooses not to ask the girl for a date; the label associated with *Night Club?* specifies a logical-or and is therefore satisfied by the state configuration *Date?=n*. If Joe chooses to go to a night club the chance variables *Meet Friends* and *Likes Club* influence Joe's comfort which in turn influences the pleasure of going to that club; the chance variable *Liking* is excluded from the decision problem since *Liking* is only included if *Club?=n* (rule (i)). Notice that this property could also be modelled by introducing a redundant state in the variable *Liking*. However, having redundant states tends to obscure the asymmetric structure of the decision problem, and is in general computationally demanding.

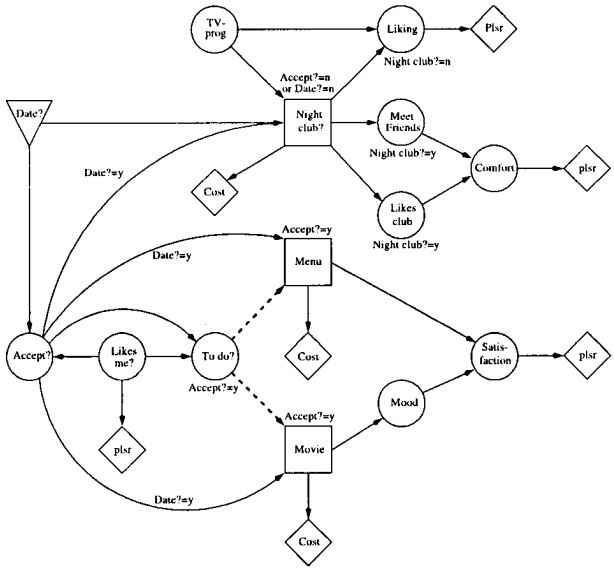

Figure 2: An AID representing "the dating problem".

Since *Date?=n* the informational arcs from *Accept?* are excluded (rule (ii)) meaning that her potential response is never observed; as previously mentioned, *Accept?* is still part of the decision problem as opposed to e.g. *Liking* if *Club?=y*. Now, as *Accept?* is never observed the variables only labeled by the state of *Accept?* are removed (rule (i)), together with all their successors (rule (iii)). The resulting decision problem is depicted in Figure 3; the variables *Accept* and *Likes me?* are included in the figure for the purpose of the AID being a tool for communication.

If Joe on the other hand chooses to ask the girl for a date he will observe her response (*Accept?*). If she declines the invitation he can choose either to go to a night club or stay at home. If she accepts the invitation the variable *To do?* will be observed which can restrict the possible decision options for *Menu* and *Movie*.

There is a small technical problem with the variable

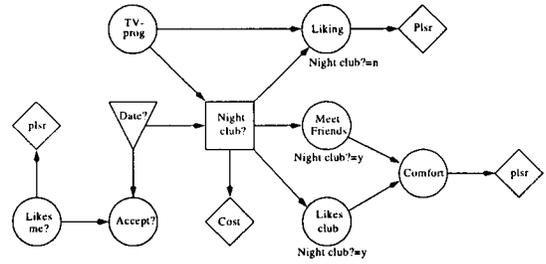

Figure 3: The dating problem if *Date?=n*.

*Satisfaction* which has the mutually exclusive variables *Menu* and *Movie* as predecessors. As no descendant of an excluded variable can be included (rule (iii)), we would unintentionally exclude *Satisfaction* whenever *Menu* or *Movie* are included. This problem can be solved by duplicating *Satisfaction* and its descendants or by adding an extra state (*no-decision*) to the variables *Menu* and *Movie*. In order to minimize redundancy in the representation we have chosen the latter.

□

### 3.1 THE QUALITATIVE LEVEL

Now, from rule (iii) we can infer the following syntactical simplification: If there exists a directed path from a node X to a node Y in $\bar{I}$, then Y "inherits" the label associated with X, i.e., Y is "effectively" labeled with $f_X \wedge f_Y$, where $f_X$ and $f_Y$ are the labels explicitly associated with X and Y, respectively. This means that we need not explicitly associate Y with the label $f_X \wedge f_Y$ (see Figure 4a). The set of variables from which a chance variable X "inherits" labels is given by $dep(X) = \mathcal{Y}$, where $\mathcal{Y}$ is the set of variables from which there exists a directed path to X in $\bar{I}$; to ensure consistency it should be noted that decision nodes, value nodes and informational arcs "inherit" labels from the empty set. E.g. in Figure 2 the variable *Meet Friends* is "effectively" conditioned on (*Night club?=y* ∧ (*Accept?=n* ∨ *Date?=n*)) since $dep(\text{Meet Friends}) = \text{Night club?}$.

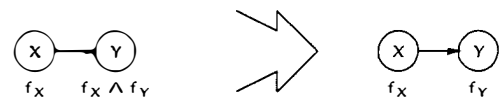

Figure 4: The figure illustrates the use of labels when specifying asymmetry.

The AID allows the specification of directed cycles with the restriction that before any of the nodes in a cycle are observed the cycle must be "broken", i.e., no matter the variables observed there must exist at least one unsatisfied label associated with a node or an informational arc in the cycle. This syntactical constraint compensates for the traditional constraint that the graph should be acyclic. Having cycles in an AID supports the specification of decisions for which the temporal order is dependent on previous observations and decisions. For instance, in Figure 5 decision $D_2$ is taken before $D_1$ if $X = x$ but $D_1$ is taken before $D_2$ if $X \neq x$.

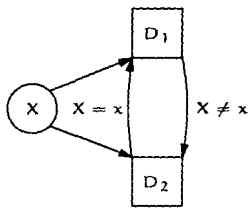

Figure 5: The figure illustrates the use of cycles in the AID.

Formally, a label is a *Boolean function* defined as a combination of *Boolean variables*, the constants *true* (1) and *false* (0) and the operators $\wedge$ (*conjunction*), $\vee$ (*disjunction*), $\neg$ (*negation*), $\Rightarrow$ (*implication*) and $\Leftrightarrow$ (*bi-implication*).

The Boolean variables are used to represent the conditioning on states e.g. if a node Y is conditioned on $X = x$ then $X = x$ should be represented as a Boolean variable in the label associated with Y. However, for ease of notation we shall use e.g. $X = x$ directly in the label (without creating an actual Boolean variable); a Boolean variable in the context of labels must therefore denote a state configuration of some node in the AID.

A *truth assignment* to a Boolean function f is the same as fixing a set of variables in the domain of f, i.e., if $X = x$ represents a Boolean variable in the domain of f, then $X = x$ can be assigned either *true* or *false* by associating X with some state $x' \in W_X$ (denoted $f[X \mapsto x']$). E.g. $(X = x)[X \mapsto x'] \equiv 1$ if $x = x'$ and $(X = x)[X \mapsto x'] \equiv 0$ otherwise.

In the remainder of this paper, we assume that each node X is associated with a label $f_X$, i.e., if X is not associated with a label in $I = (\mathcal{U}, \mathcal{E}, \mathcal{F})$, then we extend $\mathcal{F}$ with the label $f_X \equiv 1$. Moreover, we will use dom(f) to denote the domain of the label f; the domain of a label is the set of nodes referenced by the label.

A chance variable X is said to be *present* in I given configuration $\bar{c}$ over a set of variables $\mathcal{C}$, if $(f_X \wedge \bigwedge_{Y \in dep(X)} f_Y)[\mathcal{C} \mapsto \bar{c}] \equiv 1$ in I given $\bar{c}$. A chance variable X is said to be *unresolved* in I given a configuration $\bar{c}$ over a set of variables $\mathcal{C}$ if $dom(f_X[\mathcal{C} \mapsto \bar{c}]) \neq \emptyset$ $f_X[\mathcal{C} \mapsto \bar{c}] \not\equiv i(\forall i \in \{0, 1\})$, or $\exists Y \in dep(X)$ s.t. $i(f_Y[\mathcal{C} \mapsto \bar{c}]) \neq \emptyset$ and $f_Y[\mathcal{C} \mapsto \bar{c}] \not\equiv i(\forall i \in \{0, 1\})$.

The concepts *present* and *unresolved* are similarly defined for value nodes, decision nodes and informational arcs.

## 3.2 THE QUANTITATIVE LEVEL

A decision variable D is associated with a set of *restrictive functions*; a restrictive function is given by $\gamma_D : W_{\Pi_D^r} \hookrightarrow 2^{W_D}$ and specifies the legitimate decision options for D given a configuration of $\Pi_D^r \subseteq \Pi_D^R$, where $\Pi_D^R$ denotes the set of variables which can restrict the legitimate decision options for D. In Figure 2 the restrictive function associated with *Movie* specifies that the state *no-decision* is the only legitimate decision option if *To do?=restaurant* (similar for *Menu* if *To do?=movie*).

The uncertainty associated with a chance variable X is represented by a *partial conditional probability potential* $\phi_X = P(X || \Pi'_X) : W_{X \cup \Pi'_X} \hookrightarrow [0; 1]$, where $\Pi'_X = \Pi_X \setminus \mathcal{U}_T$; by definition, a test-decision variable has no probabilistic influence on a chance variable. A partial probability potential can specify that given a configuration of the conditioning set for a chance variable X, some states of X are impossible (denoted $\bot$); we make a conceptual distinction between an impossible state and a state having zero probability. Note that if all the states of X are impossible for some configuration of the conditioning set, then this must be reflected in the labeling of X e.g. if $P(X|Y)$ is only defined for $Y = y$, then $Y = y$ must occur in the label of X.

A value node X is associated with a *partial utility potential* $\psi_X : W_{\Pi_X} \hookrightarrow \mathbb{R}^+ \cup \{0\}$; requiring that the partial utility potential does not take on negative values is not an actual restriction as any utility potential can be transformed s.t. it adheres to this assumption. Furthermore, as for the PID we assume that the total utility is the sum of the local utilities.

The combination of partial potentials (addition, multiplication and division) is defined similarly to the combination of total functions by treating the undefined value ($\bot$) as an additive identity and a multiplicative zero. In particular, this ensures consistency when defining the total utility as being the sum of the local utilities.

As for the PID we define a realization of an AID as an attachment of probability and utility potentials to the appropriate variables. The probability and utility potentials associated with an AID I is denoted $\Phi_I$ and $\Psi_I$, respectively. Given a realization $\Phi_I \cup \Psi_I$ of an AID I the set of probability potentials with $X \in \mathcal{X}$ in the domain will be denoted $\Phi_{\mathcal{X}}$, i.e., $\Phi_{\mathcal{X}} = \{\phi \in \Phi_I | \exists X \in \mathcal{X} : X \in dom(\phi)\}$.

## 3.3 SPLIT CONFIGURATIONS AND DECISIONS IN CONTEXT

When taking a decision in an asymmetric decision problem, previous observations and decisions may determine the variables observed before the decision in question. For both semantic and computational reasons it is important to identify the variables actually observed before taking a particular decision. For instance, in the AID depicted in Figure 2 it would not be meaningful to have an optimal strategy for *Menu* conditioned on both *Date?=n* and *Accept?=y* since this is an impossible state configuration. So, in order to reason about the different informational states when taking a decision D we must associate D with a context describing the variables observed. That is, we need to identify the possible temporal orderings of the variables and in particular, the variables which influence the occurrence of future variables.

**Definition 4.** Let I be an AID and let $\mathcal{F}$ be the set of labels associated with I. A variable X is said to be a *split variable* in I if there exists a label $f \in \mathcal{F}$ s.t. $X \in \text{dom}(f)$. The set of split variables in I is denoted $\mathcal{S}_I$.

Now, the partial order $\prec$ induced by an AID I is found by initially treating I as a PID (ignoring any labels) and then refining the partial order $\prec'$ induced by the PID s.t. for any pair of variables X and Y, where $X \not\prec' Y$ and $Y \not\prec' X$ we have $X \prec Y$ if $X \in \mathcal{S}_I$ (see Figure 6); if $\{X, Y\} \subseteq \mathcal{S}_I$ we have $X \not\prec Y$ and $Y \not\prec X$.

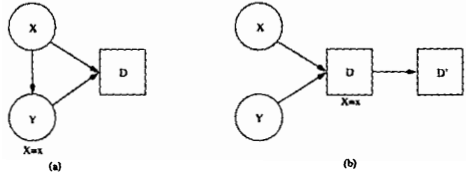

Figure 6: In figure (a) the AID induces the partial order $X \prec Y \prec D$, and in figure (b) the AID induces the partial order $X \prec Y \prec D \prec D'$.

In what follows we assume that an AID has exactly one split variable $S_\nabla^1$ (termed the *initial split variable*) satisfying that $\forall Y \in \mathcal{S}_I \setminus \{S_\nabla^1\} : S_\nabla^1 \prec Y$; *Date?* is the initial split variable in the AID depicted in Figure 2. Obviously, the set of variables succeeding the initial split variable $S_\nabla^1$ is dependent of the state of $S_\nabla^1$ hence, we define the concept of a missing variable.

**Definition 5.** Let I be an AID and let $S_\nabla^1$ be the initial split variable in I. The chance variable X is said to be *missing* in I given $S_\nabla^1 = s_1$ if:

i) $f_X[S_\nabla^1 \mapsto s_1] \equiv 0$ or

ii) $\exists Y \in \text{dep}(X)$ s.t. Y is missing given $S_\nabla^1 = s_1$ or

iii) $\forall S \in \mathcal{S}_I \setminus \{S_\nabla^1\} : X \prec S$ and X is unresolved given $S_\nabla^1 = s_1$.

The above definition is easily adopted to value nodes, decision nodes and informational arcs and will therefore not be described further.

The following definition specifies the AID obtained from another AID I by instantiating the initial split variable in I.

**Definition 6.** Let $I = (\mathcal{U}, \mathcal{E}, \mathcal{F})$ be an AID and let $S_\nabla^1$ be the initial split variable in I. The AID $I' = (\mathcal{U}', \mathcal{E}', \mathcal{F}')$ is said to be *myopicly reduced* from I given $S_\nabla^1 = s_1$ if:

- $\mathcal{U}' = \{X \in \mathcal{U} | X$ is not missing given $S_\nabla^1 = s_1\}$.
- $\mathcal{E}' = \{(X, Y) \in \mathcal{E} | \{X, Y\} \subseteq \mathcal{U}' \wedge (X, Y)$ is not missing in I given $S_\nabla^1 = s_1\}$.
- $\mathcal{F}' = \{f_X[S_\nabla^1 \mapsto s_1] | f_X \in \mathcal{F}$ and $X \in \mathcal{U}'\} \cup \{f_{(X,Y)}[S_\nabla^1 \mapsto s_1] | f_{(X,Y)} \in \mathcal{F}$ and $(X, Y) \in \mathcal{E}'\}$.

That is, we myopicly reduce an AID I by removing the missing nodes and the missing arcs. However, the removal of arcs might render additional nodes missing thus, for all missing nodes and arcs to be removed we need to remove them iteratively. The AID $I'$ obtained from I by iteratively removing missing nodes and arcs is said to be *reduced* from I given $S_\nabla^1 = s_1$ and is denoted $I[S_\nabla^1 \mapsto s_1]$.

Figure 3 illustrates the AID which has been reduced from the AID in Figure 2 given *Date?=n*. Notice that reducing an AID w.r.t. its initial split variable $S_\nabla^1$ is the same as instantiating $S_\nabla^1$ hence, $S_\nabla^1$ is not a split variable in $I[S_\nabla^1 \mapsto s_1]$.

In the remainder of this paper we restrict our attention to AIDs having exactly one initial split variable. This restriction also applies to AIDs which have been reduced from other AIDs that is, we do not consider AIDs which can be reduced to an AID that does not adhere to this restriction; having a unique initial split variable ensures that the reduction is unambiguous and it does not seem to exclude any natural decision problems; actually, decision trees have the same property. Furthermore, for ease of notation we shall treat restrictive variables as split variables unless stated otherwise ($\mathcal{S}_I^R$ denotes the union of $\mathcal{S}_I$ and the set of restrictive variables in I). However, we do not require the occurrence of a unique restrictive variable as the order in which the restrictive variables are instantiated is of no importance, i.e., the syntactical constraint on the split variables does not extend to the restrictive variables.

Now, based on the requirement about a unique split variable in an AID I we can identify the possible st

UNCERTAINTY IN ARTIFICIAL INTELLIGENCE PROCEEDINGS 2000										423configurations in I. These configurations are found by going in the temporal order specified by I. That is, we iteratively identify the initial split variable in the AID reduced from the AID in the previous step and assign this split variable a configuration consistent with the previous one.

The initial split variable $S^1_\nabla$ in I is identified as described previously. In general *the k'th split variable* in I w.r.t. the configuration $\bar{s} = (s_1, s_2, \ldots, s_{k-1})$, denoted $S^k_{\bar{s}_{k-1}}$, is the initial split variable in $I[S^1_\nabla \mapsto s_1][S^2_{\bar{s}_1} \mapsto s_2] \cdots [S^{k-1}_{\bar{s}_{k-2}} \mapsto s_{k-1}]$, where $S^i_{\bar{s}_{i-1}}$ is the initial split variable in I w.r.t. the configuration $\bar{s}_{i-1} = (s_1, s_2, \ldots, s_{i-1})$. Obviously, if I' has been reduced from I w.r.t. $S^1_\nabla = s_1$ then $s_1$ must be a possible state for $S^1_\nabla$, i.e., each time an AID is reduced the possible outcomes/decision options for the split variables are "updated". In what follows we let $I[S^1_{\bar{s}_{i-1}} \mapsto \bar{s}_i]$ denote the AID $I[S^1_\nabla \mapsto s_1][S^2_{\bar{s}_1} \mapsto s_2] \cdots [S^i_{\bar{s}_{i-1}} \mapsto s_i]$.

**Definition 7.** Let $\mathcal{S}'_I \subseteq \mathcal{S}^R_I$ and $\mathcal{S}''_I = \{S_1, S_2, \ldots, S_l\} \subseteq \mathcal{S}'_I$ be subsets of the split variables contained in the AID I. A configuration $\bar{s} = (s_1, s_2, \ldots, s_l)$ over the variables $\mathcal{S}''_I$ is said to be a *split configuration* for $\mathcal{S}'_I$ over $\mathcal{S}''_I$ if:

- $S_i$ is the i'th split variable in I w.r.t. the configuration $\bar{s}_{i-1} = (s_1, s_2, \ldots, s_{i-1})$ and

- S is not a split variable in $I[S^1_{\bar{s}_{i-1}} \mapsto \bar{s}_l]$, $\forall S \in \mathcal{S}'_I \setminus \mathcal{S}''_I$

If $\mathcal{S}'_I = \mathcal{S}^R_I$ then $\bar{s}$ is said to be an *exhaustive split configuration* for I over $\mathcal{S}''_I$.

For notational convenience, we will sometimes use $I[\mathcal{S}''_I \mapsto \bar{s}]$ to denote the AID reduced from the AID I w.r.t. the split configuration $\bar{s}$ over the variables $\mathcal{S}''_I$.

**Example 2.** Consider the AID depicted in Figure 2. The configuration $(Date?=y)$ is a split configuration, whereas $(Date?=y, Accept?=y, To\ do?=movie)$ is an exhaustive split configuration. The configuration $(Date?=n, Club?=n)$ is an exhaustive split configuration also since $Date?=n$ implies that $Accept?$ is never observed, thereby rendering the variables *Movie*, *Menu* and *To Do?* missing. □

Based on the notion of a split configuration we can determine the *contexts* in which a decision variable can occur. A context for a decision variable D is a configuration $\bar{s} = (\bar{\omega}, \bar{x})$, where:

- $\bar{s}$ is a split configuration for $\mathcal{S}'_I$ over the variables $\mathcal{S}''_I$ satisfying that there does not exist a split variable $S \in \mathcal{S}_{I[\mathcal{S}''_I \mapsto \bar{s}]}$ s.t. $S \prec D$ in $I[\mathcal{S}''_I \mapsto \bar{s}]$.

- $\bar{x}$ is a configuration over the restrictive variables for D in $I[\mathcal{S}''_I \mapsto \bar{s}]$.

The set of contexts in which a decision variable D can occur is denoted $\Omega_D$.

## 4 SOLVING ASYMMETRIC INFLUENCE DIAGRAMS

Solving an AID is the same as determining an optimal strategy for the decisions involved. However, as opposed to the PID we can not restrict our attention to the variables being required for the decision variable in question; the variables being required for D may vary depending on the context in which D appears. Thus we define a strategy as follows:

**Definition 8.** A *strategy* for an AID I is a set of functions $\Delta = \{\delta_D | D \in \mathcal{U}_D\}$, where $\delta_D$ is a decision function given by:

$$\delta_D : W_{\text{pred}(D)_{\bar{s}}} \to W_D, \forall \bar{s} \in \Omega_D,$$

where $\text{pred}(D)_{\bar{s}}$ is the set of variables preceding D under the partial order induced by the AID which has been reduced from I w.r.t. $\bar{s}$.

A strategy that maximizes the expected utility is termed *optimal strategy*, and a decision function $\delta_D$ that maximizes the expected utility for decision D w.r.t. each context $\bar{s} \in \Omega_D$ is termed an *optimal strategy for* D.

A well-defined AID (specified in the following section) can in principle be solved by unfolding it into a decision tree, and then use the "average-out and foldback" algorithm on that tree; the partial probability potentials specified by the realization of the AID can be seen as a model of the uncertainty associated with the chance variables (this is somewhat similar to the approach found in [Call and Miller, 1990]). However, this "brute force" approach would create an unnecessary large decision tree in case the original decision problem specifies symmetric subproblems.

### 4.1 DECOMPOSING ASYMMETRIC INFLUENCE DIAGRAMS

In this section we present an algorithm for solving AIDs. The main idea underlying the algorithm is to decompose the decision problem into a collection of symmetric subproblems organized in a tree structure, and then propagate from the leaves towards the root using existing evaluation methods to solve the "smaller" symmetric subproblems.

The decomposition is performed by reducing the AID w.r.t. the possible states of its initial split variable. This reduction is then applied iteratively to the AIDs produced in the previous step until no split variables remain. For instance, *Date?* is the initial split variable in "the dating problem" depicted in Figure 2 so by decomposing the problem w.r.t. *Date?=no* we obtain the AID depicted in Figure 3, where the initial split variable is *Night Club?*. An optimal strategy can then be found by iteratively eliminating the so-called free variables in each of the subproblems:

**Definition 9.** Let $I[S_\nabla^1 \mapsto s_1]$ be the AID reduced from the AID I. The variable X is said to be *free* in $I[S_\nabla^1 \mapsto s_1]$ if $S_\nabla^1 \prec X$ and $\forall S \in S_{I[S_\nabla^1 \mapsto s_1]} : X \prec S$, where $\prec$ is the partial order induced by $I[S_\nabla^1 \mapsto s_1]$. If X is not free in $I[S_\nabla^1 \mapsto s_1]$ then X is said to be *bound* in $I[S_\nabla^1 \mapsto s_1]$.

The evaluation of an AID I is initiated by invoking the algorithm *Evaluation* on I; note that in the following algorithms we exploit that instantiating the initial split variable in an AID produces another AID with a unique initial split variable.

**Algorithm 1 (Evaluation).** *Let* I *be an AID and let* $S_\nabla^1$ *be the initial split variable in* I. *If* Evaluation *is invoked on* I, *then*

i) *Invoke* Evaluation *on* $I[S_\nabla^1 \mapsto s_1], \forall s_1 \in W_{S_\nabla^1}$.

ii) *Absorb the potentials from* $I[S_\nabla^1 \mapsto s_1]$ *to* I, $\forall s_1 \in W_{S_\nabla^1}$.

iii) *Let* $\psi$ *be a utility potential absorbed (Algorithm 2) from* $I[S_\nabla^1 \mapsto s_1^i]$ *s.t.* $S_\nabla^1 \notin dom(\psi)$. *If* $\exists j \neq i$ *s.t* $\psi$ *is not absorbed from* $I[S_\nabla^1 \mapsto S_1^j]$ *to* I *then condition* $\psi$ *on* $S = s_1^i$ *(see Figure 7)*.

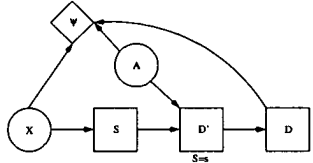

Figure 7: The occurrence of D' is dependent on the state of S thus, the utility potential produced by the elimination of D, A and D' is dependent on S; the relative temporal order of A and D vary w.r.t. the state of S.

**Algorithm 2 (Absorption).** *Let* $I[S_{\bar{s}_{i-1}}^i \mapsto \bar{s}_i]$ *be an AID and let* S *be the initial split variable in* $I[S_{\bar{s}_{i-1}}^i \mapsto \bar{s}_i]$. *If* Absorption *is invoked on* $I[S_{\bar{s}_{i-1}}^i \mapsto \bar{s}_i][S \mapsto s]$ *from* $I[S_{\bar{s}_{i-1}}^i \mapsto \bar{s}_i]$, *then:*

i) *Let* $\mathcal{X}$ *be the free variables* $I[S_{\bar{s}_{i-1}}^i \mapsto \bar{s}_i][S \mapsto s]$ *and set* $\Lambda = \{\lambda \in \Phi_{I[S_{\bar{s}_{i-1}}^i \mapsto \bar{s}_i][S \mapsto s]} \cup \Psi_{I[S_{\bar{s}_{i-1}}^i \mapsto \bar{s}_i][S \mapsto s]} | \exists X \in \mathcal{X} \text{ s.t. } X \in dom(\lambda)\}$.

ii) *Eliminate the variables* $\mathcal{X}$ *from* $\Lambda$ *w.r.t. the partial order induced by* $I[S_{\bar{s}_{i-1}}^i \mapsto \bar{s}_i][S \mapsto s]$ *(Algorithm 3). Let* $\Phi_{I[S \mapsto s]}^*$ *and* $\Psi_{I[S \mapsto s]}^*$ *be the sets of probability potentials and utility potentials obtained.*

iii) *For each* $\lambda \in \Lambda$, *remove* $\lambda$ *from* $I[S_{\bar{s}_{j-1}}^j \mapsto \bar{s}_j]$ *and associate* $\Phi_{I[S \mapsto s]}^* \cup \Psi_{I[S \mapsto s]}^*$ *with* $I[S_{\bar{s}_{j-1}}^j \mapsto \bar{s}_j]$, *for all* $1 \leq j \leq i$.

The algorithm below describes the elimination of variables, and is inspired by the *lazy evaluation* architecture[Madsen and Jensen, 1999]. However, any elimination algorithm can in principle be used.

**Algorithm 3 (Elimination).** *Let* I *be an AID and let* $\Phi_I$ *and* $\Psi_I$ *be the sets of probability and utility potentials associated with* I. *If* Elimination *of variable* X *is invoked on* $\Phi_I \cup \Psi_I$, *then:*

i) *Set* $\Phi_X = \{\phi \in \Phi_I | X \in dom(\phi)\}$ *and* $\Psi_X = \{\psi \in \Psi_I | X \in dom(\psi)\}$.

ii) *Calculate:*

$$\phi_X^* = \mathsf{M}_X \prod_{\phi \in \Phi_X} \phi$$

$$\psi_X^* = \mathsf{M}_X \prod_{\phi \in \Phi_X} \phi \sum_{\psi \in \Psi_X} \psi,$$

*where* $\mathsf{M}$ *is a marginalization operator depending on the type of* X, *i.e.,* $\mathsf{M}$ *denotes a summation if* X *is a chance variable and a maximization if* X *is a decision variable.*

iii) *Return* $\Phi_I^* = \Phi_I \backslash \Phi_X \cup \{\phi_X^*\}$ *and* $\Psi_I^* = \Psi_I \backslash \Psi_X \cup \{\frac{\psi_X^*}{\phi_X^*}\}$

During the evaluation, the decision option maximizing the utility potential from which a decision variable D is eliminated should be recorded as the optimal strategy for D w.r.t. to the context in question.

Now, based on the algorithms above we define the concept of a well-defined AID. The definition is based on the notion of a significant chance variable, which can be adopted from the PID framework by considering the admissible total orderings of the free variables in each of the AIDs produced during the decomposition; the structural characterization of the chance variables being significant for a given decision variable (see [Nielsen and Jensen, 1999b]) can also be adopted to an AID I, except that we have to investigate each of the AIDs reduced from I w.r.t. the exhaustive split configurations for I.

**Definition 10 (Well-defined).** An AID I is said to *define a decision scenario* if:

- for all split configurations $\bar{s}$, there does not exist a free chance variable A and a free decision variable

D in $I[\mathcal{S}_I'' \mapsto \bar{s}]$ s.t. A is significant for D; $\bar{s}$ is a configuration over the variables $\mathcal{S}_I'' \subseteq \mathcal{S}_I^R$

- for any decision variable D and for each context $\bar{s} = (\bar{\omega}, \bar{x})$ for D there does not exist two restrictive functions $\gamma_D^1$ and $\gamma_D^2$ s.t. $\text{dom}(\gamma_D^1) \subseteq \Pi_{D|\bar{s}}^R$ and $\text{dom}(\gamma_D^2) \subseteq \Pi_{D|\bar{s}}^R$, where $\Pi_{D|\bar{s}}^R$ is the set of restrictive variables which are present in I given $\bar{s}$.

**Theorem 2 (Sound).** If Algorithm 1 is invoked on an AID I which define a decision scenario, then Algorithm 1 computes an optimal strategy for each decision variable in I.

*Proof.* The idea of the proof is to initially treat non-split variables as split variables, thereby obtaining a decision tree representation of the decision problem when reducing the AID; each subproblem contains exactly one free variable which corresponds to a node in the decision tree. Note that from Algorithm 1 we have that treating non-split variables as split variables has no impact on the evaluation.

Finally, we exploit that the set of partial probability potentials constitutes a model of the uncertainty associated with the chance variables, and from this it can be shown that the calculations performed by Algorithm 1 are equivalent to the calculations peformed when solving the corresponding decision tree. For further detail see [Nielsen and Jensen, 1999a]. □

## 5 CONCLUSION

In this paper we have presented a framework, termed asymmetric influence diagrams, for representing asymmetric decision problems. The asymmetric influence diagram is based on the partial influence diagram and uses labels, associated with nodes and informational arcs, to encode structural asymmetry at the qualitative level. Asymmetry which deals with the possible outcomes of an observation or the legitimate decision options of a decision variable is represented in partial probability potentials and restrictive functions, respectively.

We have presented an algorithm for solving asymmetric influence diagrams. The algorithm decomposes the asymmetric decision problem into a collection of symmetric subproblems which can be solved using existing methods for solving influence diagrams.

As part of the future work, the class of asymmetric decision problems which can be modeled effectively using AIDs needs to be determined. We claim that the language of AIDs is as strong as that of decision trees, but the amount of redundancy in the models should be determined.